# Comment on "robustness and regularization of support vector machines" by H. Xu, et al., (Journal of Machine Learning Research, vol. 10, pp. 1485-1510, 2009)

**Yahya Forghani**  YAHYAFOR2000@YAHOO.COM
Department of Computer,
Ferdowsi University of Mashhad,
Islamic Azad University, Mashhad Branch
Mashhad, IRAN

**Hadi Sadoghi Yazdi**  H-SADOGHI@UM.AC.IR
Department of Computer,
Ferdowsi University of Mashhad,
Center of Excellence on Soft Computing and Intelligent Information Processing,
Mashhad, IRAN

## Abstract

This paper comments on the published work dealing with robustness and regularization of support vector machines (Journal of Machine Learning Research, vol. 10, pp. 1485-1510, 2009) by H. Xu, etc. They proposed a theorem to show that it is possible to relate robustness in the feature space and robustness in the sample space directly. In this paper, we propose a counter example that rejects their theorem.

## 1. Comment

Firstly, it must be stated that H. Xu, et al. made a good study of robustness and regularization of support vector machines in (Xu, et al. 2009). They proposed the following theorem to show that it is possible to relate robustness in the feature space and robustness in the sample space directly:

**Theorem 14 [1]**. *Suppose that the kernel function has the form* $k(x, x') = f(\|x - x'\|)$, *with* $f: R^+ \to R$ *a decreasing function. Denote by* $\mathcal{H}$ *the RKHS space of* $k(.,.)$ *and* $\Phi(.)$ *the corresponding feature mapping. Then we have any* $x \in R^n$, $w \in \mathcal{H}$ *and* $c > 0$,

$$\sup_{\|\delta\| \leq c} \langle w, \Phi(x - \delta) \rangle = \sup_{\|\delta_\phi\|_{\mathcal{H}} \leq \sqrt{2f(0) - 2f(c)}} \langle w, \Phi(x) - \delta_\phi \rangle.$$

The following counter example rejects the mentioned theorem. However, this theorem is a standalone result in the appendix of (Xu, et al., 2009), which is not used anywhere else in (Xu, et al., 2009). Thus, the main result and all other results of (Xu, et al., 2009) are not affected in any way.

**Counter example.** Let $\Phi(.)$ be the feature mapping of Gaussian kernel function. We have $\|\Phi(x)\| = 1$. Let $w = \Phi(x)$. Therefore, $\langle w, \Phi(x) \rangle = \|w\|$, and
$$\sup_{\|\delta\| \leq c} \langle w, \Phi(x - \delta) \rangle = \|w\|. \tag{1}$$
Moreover,
$$\sup_{\|\delta_\phi\|_{\mathcal{H}} \leq \sqrt{2f(0) - 2f(c)}} \langle w, \Phi(x) - \delta_\phi \rangle =$$
$$\sup_{\|\delta_\phi\|_{\mathcal{H}} \leq \sqrt{2f(0) - 2f(c)}} \langle w, \Phi(x) \rangle + \sup_{\|\delta_\phi\|_{\mathcal{H}} \leq \sqrt{2f(0) - 2f(c)}} \langle w, \delta_\phi \rangle =$$
$$\|w\| + \sup_{\|\delta_\phi\|_{\mathcal{H}} \leq \sqrt{2f(0) - 2f(c)}} \langle w, \delta_\phi \rangle =$$
$$\|w\| + \|w\|\sqrt{2f(0) - 2f(c)}. \tag{2}$$
According to Eq. (1) and Eq. (2), and since f is a decreasing function, for any $c > 0$, we have
$$\sup_{\|\delta\| \leq c} \langle w, \Phi(x - \delta) \rangle < \sup_{\|\delta_\phi\|_{\mathcal{H}} \leq \sqrt{2f(0) - 2f(c)}} \langle w, \Phi(x) - \delta_\phi \rangle.$$
End of counter example.

The exact spot that the error has been occurred in the mentioned theorem is Eq. (19) of (Xu, et al., 2009). There it has been claimed that the image of the RKHS feature mapping is dense, which unfortunately is not true. Indeed, because $\langle \Phi(x), \Phi(x) \rangle = K(0)$ where $K(.)$ is the kernel function, the image of the feature mapping is in a ball of radius $\text{sqrt}(K(0))$.